\nonstopmode
\documentclass{article} 
\usepackage{iclr2017_conference,times}
\usepackage{amsfonts}
\usepackage{amsmath}
\usepackage{bbm}
\usepackage{booktabs}
\usepackage[T1]{fontenc}
\usepackage{hyperref}
\usepackage{mathtools}
\usepackage{pgf}
\usepackage{rotating}
\usepackage{url}

\title{Latent Sequence Decompositions}

\author{William Chan\thanks{Work done at Google Brain.} \\
  Carnegie Mellon University \\
  \texttt{williamchan@cmu.edu} \\
  \AND
  Yu Zhang\footnotemark[1] \\
  Massachusetts Institute of Technology \\
  \texttt{yzhang87@mit.edu}
  \AND
  Quoc V. Le, Navdeep Jaitly \\
  Google Brain \\
  \texttt{\{qvl,ndjaitly\}@google.com}
}

\iclrfinalcopy 

\begin{document}

\maketitle

\begin{abstract}
Sequence-to-sequence models rely on a fixed decomposition of the target sequences into
a sequence of tokens that may be words, word-pieces or characters. The choice of these tokens
and the decomposition of the target sequences into a sequence of tokens is often static,
and independent of the input, output data domains. This can potentially lead
to a sub-optimal choice of token dictionaries, as the decomposition is not informed
by the particular problem being solved. In this paper we present Latent Sequence
Decompositions (LSD), a framework in which the decomposition of sequences into
constituent tokens is learnt during the training of the model. The decomposition
depends both on the input sequence and on the output sequence. In LSD,
during training, the model samples decompositions incrementally, from
left to right by locally sampling between valid extensions. We experiment with
the Wall Street Journal speech recognition task. Our LSD model achieves 12.9\%
WER compared to a character baseline of 14.8\% WER. When combined with a
convolutional network on the encoder, we achieve a WER of 9.6\%.
\end{abstract}

\section{Introduction}
Sequence-to-sequence (seq2seq) models
\citep{sutskever-nips-2014,cho-emnlp-2014} with attention
\citep{bahdanau-iclr-2015} have been successfully applied to many applications
including machine translation \citep{luong-acl-2015,jean-acl-2015}, parsing
\citep{vinyals-nips-2015}, image captioning
\citep{vinyals-cvpr-2014,xu-icml-2015} and Automatic Speech Recognition (ASR)
\citep{chan-icassp-2016,bahdanau-icassp-2016}.

Previous work has assumed a fixed deterministic decomposition for each output
sequence.  The output representation is usually a fixed sequence of words
\citep{sutskever-nips-2014,cho-emnlp-2014}, phonemes
\citep{chorowski-nips-2015}, characters
\citep{chan-icassp-2016,bahdanau-icassp-2016} or even a mixture of characters
and words \citep{thang-acl-2016}. However, in all these cases, the models are
trained towards one fixed decomposition for each output sequence.

We argue against using fixed deterministic decompositions of a sequence that
has been defined a priori. Word segmented models \citep{luong-acl-2015,jean-acl-2015} often have
to deal with large softmax sizes, rare words and Out-of-Vocabulary (OOV)
words. Character models \citep{chan-icassp-2016,bahdanau-icassp-2016} overcome
the OOV problem by modelling the smallest output unit, however this typically
results in long decoder lengths and computationally expensive inference. And even with
mixed (but fixed) character-word models \citep{thang-acl-2016}, it is unclear
whether such a predefined segmentation is optimal. In all these examples, the
output decomposition is only a function of the output sequence. This may be
acceptable for problems such as translations, but inappropriate for tasks
such as speech recognition, where segmentation should also be informed by
the characteristics of the inputs, such as audio.

We want our model to have the capacity and flexibility to learn a distribution
of sequence decompositions. Additionally, the decomposition should be a
sequence of variable length tokens as deemed most probable.  For example,
language may be more naturally represented as word pieces
\citep{schuster-icassp-2012} rather than individual characters. In many speech
and language tasks, it is probably more efficient to model ``qu'' as one output
unit rather than ``q'' + ``u'' as separate output units (since in English,
``q'' is almost always followed by ``u''). Word piece models also naturally
solve rare word and OOV problems similar to character models.

The output sequence decomposition should be a function of both the input
sequence and the output sequence (rather than output sequence alone).  For
example, in speech, the choice of emitting ``ing'' as one word piece or as
separate tokens of ``i'' + ``n'' + ``g'' should be a function of the current
output word as well as the audio signal (i.e., speaking style).

We present the Latent Sequence Decompositions (LSD) framework.  LSD does not
assume a fixed decomposition for an output sequence, but rather learns to
decompose sequences as function of both the input and the output sequence. Each
output sequence can be decomposed to a set of latent sequence decompositions
using a dictionary of variable length output tokens. The LSD framework produces
a distribution over the latent sequence decompositions and marginalizes over them
during training.  During test inference, we find the best decomposition and
output sequence, by using beam search to find the most likely output sequence
from the model.

\section{Latent Sequence Decompositions}
\label{sec:lsd}
In this section, we describe LSD more formally.
Let $\mathbf x$ be our input sequence, $\mathbf y$ be our output sequence and
$\mathbf z$ be a latent sequence decomposition of $\mathbf y$. The latent
sequence decomposition $\mathbf z$ consists of a sequence of $z_i \in
\mathcal{Z}$ where $\mathcal{Z}$ is the constructed token space.
Each token $z_i$ need not be the same length, but rather in our framework, we expect the
tokens to have different lengths. Specifically, $\mathcal Z \subseteq \cup_{i=1}^n \mathcal
C^i$ where $\mathcal C$  is the set of singleton tokens and $n$ is the length
of the largest output token. In ASR , $\mathcal C$ would
typically be the set of English characters, while $\mathcal Z$ would be word pieces
(i.e., $n$-grams of characters).

To give a concrete example, consider a set of tokens
$\left\{\textrm{``a''}, \textrm{``b''}, \textrm{``c''}, \textrm{``at''}, \textrm{``ca''}, \textrm{``cat''}\right\}$.  With this set of tokens,
the word ``cat'' may be represented as the sequence ``c'', ``a'', ``t'',
or the sequence ``ca'', ``t'', or alternatively as the single token ``cat''. Since
the appropriate decomposition of the word ``cat'' is not known a priori, the decomposition
itself is latent.

Note that the length $|\mathbf z_a|$ of a decomposition $\mathbf z_a$ need not be
the same as the length of the output sequence, $|\mathbf y|$ (for example ``ca'', ``t'' has a
length of 2, whereas the sequence is 3 characters long). Similarly, a different
decomposition $\mathbf z_b$ (for example the 3-gram token ``cat'') of the same sequence
may be of a different length (in this case 1).

Each decomposition, collapses to the target output sequence using a trivial
collapsing function $\mathbf y = \mathrm{collapse}(\mathbf z)$. Clearly, the
set of decompositions,  $\{\mathbf z : \mathrm{collapse}(\mathbf z) = \mathbf y\}$,
of a sequence, $\mathbf y$, using a non-trivial token set, $\mathcal Z$,
can be combinatorially large.

If there was a known, unique, correct segmentation  $\mathbf z^*$ for a given
pair, $(\mathbf x, \mathbf y)$, one could simply train the model to output
the fixed deterministic decomposition $\mathbf z^*$. However, in most problems,
we do not know the best possible decomposition $\mathbf z^*$; indeed it may be
possible that the output can be correctly decomposed into multiple alternative
but valid segmentations.  For example, in end-to-end ASR we typically use
characters as the output unit of choice \citep{chan-icassp-2016, bahdanau-icassp-2016} but
word pieces may be better units as they more closely align with the
acoustic entities such as syllables. However, the most appropriate decomposition
$\mathbf z^*$ for a given is $(\mathbf x, \mathbf y)$ pair is often unknown. Given
a particular $\mathbf y$, the best $\mathbf z^*$ could even change depending
on the input sequence $\mathbf x$ (i.e., speaking style).

In LSD, we want to learn a probabilistic segmentation mapping from $\mathbf x
\rightarrow \mathbf z \rightarrow \mathbf y$.  The model produces a
distribution of decompositions, $\mathbf z$, given an input sequence $\mathbf x$,
and the objective is to maximize the log-likelihood of the ground truth
sequence $\mathbf y$. We can accomplish this by factorizing and marginalizing
over all possible $\mathbf z$ latent sequence decompositions under our model
$p(\mathbf z | \mathbf x; \theta)$ with parameters $\theta$:
\begin{align}
	\log p(\mathbf y | \mathbf x; \theta) &= \log \sum_{\mathbf z} p(\mathbf y, \mathbf z | \mathbf x; \theta) \\
	&= \log \sum_{\mathbf z} p(\mathbf y | \mathbf z, \mathbf x) p(\mathbf z | \mathbf x; \theta) \\
	&= \log \sum_{\mathbf z} p(\mathbf y | \mathbf z) p(\mathbf z | \mathbf x; \theta)
\end{align}
where $p(\mathbf y | \mathbf z) = \mathbbm{1}(\mathrm{collapse}(\mathbf z) =
\mathbf y)$ captures path decompositions $\mathbf z$ that collapses to
$\mathbf y$. Due to the exponential number of decompositions of $\mathbf y$,
exact inference and search is intractable for any non-trivial token set $\mathcal Z$
and sequence length $|\mathbf y|$. We describe a beam search algorithm to do
approximate inference decoding in Section \ref{sec:decoding}.

Similarly, computing the exact gradient is intractable. However, we can derive
a gradient estimator by differentiating w.r.t. to $\theta$ and taking its
expectation:
\begin{align}
	\frac{\partial}{\partial \theta} \log p(\mathbf y | \mathbf x; \theta) &= \frac{1}{p(\mathbf y | \mathbf x; \theta)} \frac{\partial}{\partial \theta} \sum_z p(\mathbf y | \mathbf x, \mathbf z) p(\mathbf z | \mathbf x; \theta) \\
	&= \frac{1}{p(\mathbf y | \mathbf  x; \theta)} \sum_z p(\mathbf y | \mathbf x, \mathbf z) \nabla_\theta p(\mathbf z | \mathbf x; \theta) \\
 &= \frac{1}{p(\mathbf y | \mathbf  x; \theta)} \sum_z p(\mathbf y | \mathbf x, \mathbf z) p(\mathbf z | \mathbf x; \theta) \nabla_\theta \log p(\mathbf z | \mathbf x; \theta) \label{eqn:logtrick} \\
 &= \mathbb{E}_{\mathbf z \sim p(\mathbf z | \mathbf x, \mathbf y; \theta)} \left[\nabla_\theta \log p(\mathbf z | \mathbf x; \theta)\right] \label{eqn:gradient-estimator}
\end{align}

Equation \ref{eqn:logtrick} uses the identity $\nabla_\theta f_\theta(x) = f_\theta(x) \nabla_\theta \log f_\theta(x)$ assuming $f_\theta(x) \neq 0 \; \forall \; x$.
Equation \ref{eqn:gradient-estimator} gives us an unbiased estimator of our
gradient. It tells us to sample some latent sequence decomposition $\mathbf z
\sim p(\mathbf z | \mathbf y, \mathbf x; \theta)$ under our model's posterior, where $\mathbf z$ is constraint to be
a valid sequence that collapses to $\mathbf y$, i.e.
$\mathbf z \in \{\mathbf z' : \mathrm{collapse}(\mathbf z') = \mathbf y\}$.
To train the model, we sample $\mathbf z \sim p(\mathbf z | \mathbf y, \mathbf x; \theta)$ and compute the gradient
of $\nabla_\theta \log p(\mathbf z | \mathbf x; \theta)$ using backpropagation.
However, sampling $\mathbf z \sim p(\mathbf z | \mathbf y, \mathbf x; \theta)$
is difficult. Doing this exactly is computationally expensive, because it would
require sampling correctly from the posterior -- it would be possible to do
this using a particle filtering like algorithm, but would require a full
forward pass through the output sequence to do this.

Instead, in our implementation we use a heuristic to sample $\mathbf z
\sim p(\mathbf z | \mathbf y, \mathbf x; \theta)$. At each output time step
$t$ when producing tokens $z_1, z_2 \cdots z_{\left(t-1\right)}$, we sample from
$z_t \sim p\left(z_t | \mathbf x, \mathbf y, \mathbf z_{<t}, \theta\right)$
in a left-to-right fashion. In other words, we sample valid extensions
at each time step $t$.
At the start of the training, this left-to-right sampling procedure is not a
good approximation to the posterior, since the next step probabilities at a
time step include probabilities of all future paths from that point.

For example,
consider the case when the target word is ``cat'', and the vocabulary includes all
possible characters and the tokens ``ca'', and ``cat''. At time step 1, when the valid
next step options are ``c'', ``ca'', ``cat'', their relative probabilities reflect all possible
sequences ``c*'', ``ca*'', ``cat*'' respectively, that start from the first time
step of the model. These sets of sequences include sequences other
than the target sequence ``cat''. Thus sampling from the distribution at step 1 is a biased
procedure.

However, as training proceeds the model places more and more mass
only on the correct hypotheses, and the relative probabilities that the model
produces between valid extensions gets closer to the posterior.
In practice, we find that the when the model is trained with this method, it
quickly collapses to using single character targets, and never escapes from
this local minima\footnote{One notable exception was the word piece
``qu'' (``u'' is almost always followed by ``q'' in English). The model does
learn to consistently emit ``qu'' as one token and never produce ``q'' + ``u''
as separate tokens.}.  Thus, we follow an $\epsilon$-greedy
exploration strategy commonly found in reinforcement learning literature \citep{sutton-1998} --
we sample $z_t$ from a mixture of a uniform distribution over valid next tokens
and $p\left(z_t | \mathbf x, \mathbf y, \mathbf z_{<t}, \theta\right)$. The
relative probability of using a uniform distribution vs. $p\left( \cdot | \mathbf x,
\mathbf y, \mathbf z_{<t}, \theta\right)$ is varied over training. With
this modification the model learns to use the longer n-grams of characters
appropriately, as shown in later sections.

\section{Model}
In this work, we model the latent sequence decompositions $p(\mathbf z | \mathbf x)$ with an attention-based seq2seq model \citep{bahdanau-iclr-2015}.
Each output token $z_i$ is modelled as a conditional distribution over all
previously emitted tokens $\mathbf z_{<i}$ and the input sequence $\mathbf x$ using the chain rule:
\begin{align}
	p(\mathbf z | \mathbf x; \theta) = \prod_i p(z_i | \mathbf x, \mathbf z_{<i})
\end{align}

The input sequence $\mathbf x$ is processed through an $\mathrm{EncodeRNN}$ network. The $\mathrm{EncodeRNN}$ function transforms the features $\mathbf x$ into some higher level representation $\mathbf h$. In our experimental implementation $\mathrm{EncodeRNN}$ is a stacked Bidirectional LSTM (BLSTM)
\citep{schuster-ieeetsp-1997,graves-asru-2013} with hierarchical subsampling
\citep{hihi-nips-1996,koutnik-icml-2014}:
\begin{align}
	\mathbf h = \mathrm{EncodeRNN}(\mathbf x)
\end{align}

The output sequence $\mathbf z$ is generated with an attention-based transducer \citep{bahdanau-iclr-2015} one $z_i$ token at a time:
\begin{align}
	s_i &= \mathrm{DecodeRNN}([z_{i - 1}, c_{i - 1}], s_{i - 1}) \\
	c_i &= \mathrm{AttentionContext}(s_i, \mathbf h) \\
	p(z_i | \mathbf{x}, \mathbf z_{<i}) &= \mathrm{TokenDistribution}(s_i, c_i)
\end{align}

The $\mathrm{DecodeRNN}$ produces a transducer state $s_i$ as a function of the previously emitted token $z_{i-1}$, previous attention context $c_{i - 1}$ and previous transducer state $s_{i - 1}$. In our implementation, $\mathrm{DecodeRNN}$ is a LSTM \citep{hochreiter-neuralcomputation-1997} function without peephole connections.

The $\mathrm{AttentionContext}$ function generates $c_i$ with a content-based MLP attention network \citep{bahdanau-iclr-2015}. Energies $e_i$ are computed as a function of the encoder features $\mathbf h$ and current transducer state $s_i$. The energies are normalized into an attention distribution $\alpha_i$. The attention context $c_i$ is created as a $\alpha_i$ weighted linear sum over $\mathbf h$:
\begin{align}
	e_{i, j} &= \langle v, \tanh(\phi(s_i, h_j)) \rangle \\
	\alpha_{i, j} &= \frac{\exp(e_{i, j})}{\sum_{j'} \exp(e_{i,j'})} \\
	c_i      &= \sum_j \alpha_{i,j} h_j
\end{align}
where $\phi$ is linear transform function. $\mathrm{TokenDistribution}$ is a MLP function with softmax outputs modelling the distribution $p(z_i | \mathbf{x}, \mathbf z_{<i})$.

\section{Decoding}
\label{sec:decoding}

During inference we want to find the most likely word sequence given
the input acoustics:
\begin{align}
  \hat{\mathbf y} = \arg \max_{\mathbf y} \sum_{\mathbf z} \log p(\mathbf y | \mathbf z) p(\mathbf z | \mathbf x)
\end{align}
however this is obviously intractable for any non-trivial token space and
sequence lengths. We simply approximate this by decoding for the best word piece sequence $\hat{\mathbf z}$ and then collapsing it to its corresponding word sequence $\hat{\mathbf y}$:
\begin{align}
  \hat{\mathbf z} &= \arg \max_{\mathbf z} \log p(\mathbf z | \mathbf x) \\
  \hat{\mathbf y} &= \mathrm{collapse}(\hat{\mathbf z})
\end{align}
We approximate for the best $\hat{\mathbf z}$ sequence by doing a left-to-right
beam search \citep{chan-icassp-2016}.

\section{Experiments}
We experimented with the Wall Street Journal (WSJ) ASR task. We used the
standard configuration of train si284 dataset for training, dev93 for
validation and eval92 for test evaluation. Our input features were 80
dimensional filterbanks computed every 10ms with delta and delta-delta
acceleration normalized with per speaker mean and variance as generated by
Kaldi \citep{povey-asru-2011}. The $\mathrm{EncodeRNN}$ function is a 3 layer
BLSTM with 256 LSTM units per-direction (or 512 total) and $4=2^2$ time factor
reduction. The $\mathrm{DecodeRNN}$ is a 1 layer LSTM with 256 LSTM units. All
the weight matrices were initialized with a uniform distribution
$\mathcal{U}(-0.075, 0.075)$ and bias vectors to $0$.  Gradient norm clipping
of $1$ was used, gaussian weight noise $\mathcal{N}(0, 0.075)$ and L2 weight
decay $1\mathrm{e}{-5}$ \citep{graves-nips-2011}. We used ADAM with the default
hyperparameters described in \citep{kingma-iclr-2015}, however we decayed the
learning rate from $1\mathrm{e}{-3}$ to $1\mathrm{e}{-4}$. We used 8 GPU
workers for asynchronous SGD under the TensorFlow framework
\citep{tensorflow2015-whitepaper}. We monitor the dev93 Word Error Rate (WER)
until convergence and report the corresponding eval92 WER.  The models took around 5
days to converge.

We created our token vocabulary $\mathcal Z$ by looking at the $n$-gram
character counts of the training dataset. We explored $n \in \{2, 3, 4, 5\}$
and took the top $\{256, 512, 1024\}$ tokens based on their count frequencies
(since taking the full $n$-cartesian exponent of the unigrams would result in
an intractable number of tokens for $n > 2$). We found very minor differences
in WER based on the vocabulary size, for our $n=\{2,3\}$ word piece experiments we used a
vocabulary size of 256 while our $n=\{4,5\}$ word piece experiments used a vocabulary size of
512. Additionally, we restrict $\langle \mathrm{space} \rangle$ to be a unigram
token and not included in any other word pieces, this forces the decompositions
to break on word boundaries.

\begin{table}[t]
  \centering
  \caption{Wall Street Journal test eval92 Word Error Rate (WER) varying the $n$ sized word piece vocabulary without any dictionary or language model. We compare Latent Sequence Decompositions (LSD) versus the Maximum Extension (MaxExt) decomposition. The LSD models all learn better decompositions compared to the baseline character model, while the MaxExt decomposition appears to be sub-optimal.}
  \label{tab:wordpiece}
  \begin{tabular}{lcc}
    \toprule
    $n$ & \bfseries LSD WER & \bfseries MaxExt WER \\
    \midrule
    Baseline & \multicolumn{2}{c}{14.76} \\
    \midrule
    2 & 13.15 & 15.56 \\
    3 & 13.08 & 15.61 \\
    4 & \textbf{12.88} & 14.96 \\
    5 & 13.52 & 15.03 \\
    \bottomrule
  \end{tabular}
\end{table}

\begin{figure}[t]
  \centering
  \input{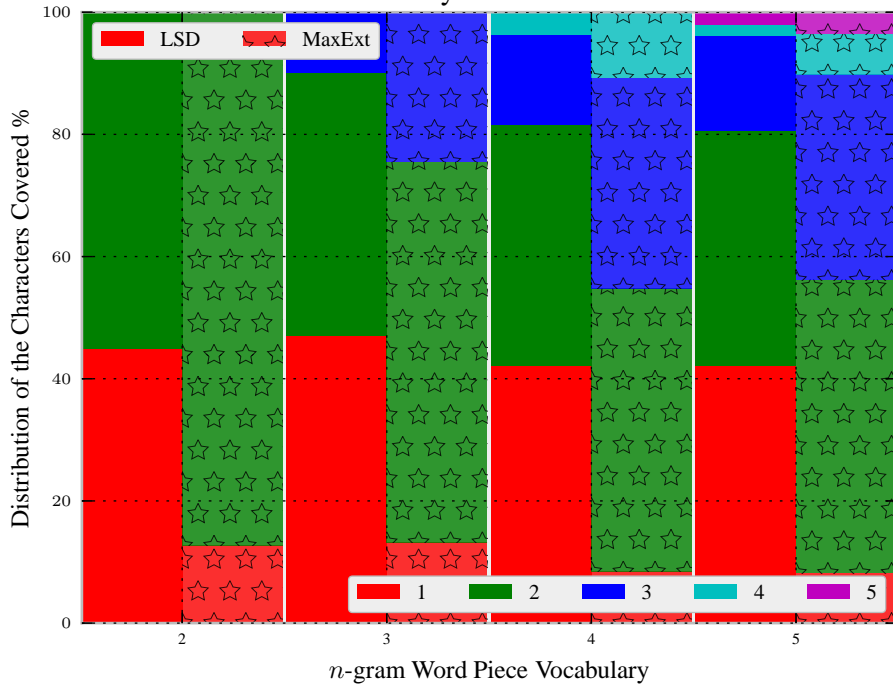}
  \caption{Distribution of the characters covered by the n-grams of the word piece models. We train Latent Sequence Decompositions (LSD) and Maximum Extension (MaxExt) models with $n \in \{2,3,4,5\}$ sized word piece vocabulary and measure the distribution of the characters covered by the word pieces. The bars with the solid fill represents the LSD models, and the bars with the star hatch fill represents the MaxExt models. Both the LSD and MaxExt models prefer to use $n \geq 2$ sized word pieces to cover the majority of the characters. The MaxExt models prefers longer word pieces to cover characters compared to the LSD models.}
  \label{fig:charactercoverage}
\end{figure}

Table \ref{tab:wordpiece} compares the effect of varying the $n$ sized word
piece vocabulary. The Latent Sequence Decompositions (LSD) models were trained
with the framework described in Section \ref{sec:lsd} 
and the (Maximum Extension) MaxExt decomposition is a fixed decomposition. MaxExt is generated in a left-to-right fashion, where at each step the longest word piece extension is selected from the vocabulary.
The MaxExt decomposition is not the shortest $|\mathbf z|$
possible sequence, however it is a deterministic decomposition that can be
easily generated in linear time on-the-fly. We decoded these models with simple
n-best list beam search without any external dictionary or Language Model (LM).

The baseline model is simply the unigram or character model and achieves 14.76\%
WER.  We find the LSD $n=4$ word piece vocabulary model to perform the best
at 12.88\% WER or yielding a 12.7\% relative improvement over the baseline
character model. None of our MaxExt models beat our character model baseline,
suggesting the maximum extension decomposition to be a poor decomposition
choice. However, all our LSD models perform better than the baseline suggesting
the LSD framework is able to learn a decomposition better than the baseline
character decomposition.

We also look at the distribution of the characters covered based on the word piece lengths during
inference across different $n$ sized word piece vocabulary used in training. We
define the distribution of the characters covered as the percentage of characters covered by the
set of word pieces with the same length across the test set, and we exclude
$\langle \mathrm{space} \rangle$ in this statistic. Figure
\ref{fig:charactercoverage} plots the distribution of the
$\{1,2,3,4,5\}$-ngram word pieces the model decides to use to decompose the
sequences.  When the model is trained to use the bigram word piece vocabulary,
we found the model to prefer bigrams (55\% of the characters emitted) over
characters (45\% of the characters emitted) in the LSD decomposition. This
suggest that a character only vocabulary may not be the best vocabulary to
learn from. Our best model, LSD with $n=4$ word piece vocabulary, covered the
word characters with 42.16\%, 39.35\%, 14.83\% and 3.66\% of the time using 1,
2, 3, 4 sized word pieces respectively.
In the $n=5$ word piece vocabulary model, the LSD model uses the $n=5$ sized word
pieces to cover approximately 2\% of the characters.  We suspect if we used a
larger dataset, we could extend the vocabulary to cover even larger $n \geq 5$.

The MaxExt model were trained to greedily emit the longest possible word piece,
consequently this prior meant the model will prefer to emit long word pieces
over characters. While this decomposition results in the shorter $|\mathbf z|$
length, the WER is slightly worse than the character baseline. This suggest the
much shorter decompositions generated by the MaxExt prior may not be best
decomposition. This falls onto the principle that the best $\mathbf z^*$
decomposition is not only a function of $\mathbf y^*$ but as a function of
$(\mathbf x, \mathbf y^*)$. In the case of ASR, the segmentation is a function
of the acoustics as well as the text.

Table \ref{tab:compare} compares our WSJ results with other published
end-to-end models. The best CTC model achieved 27.3\% WER with REINFORCE
optimization on WER \citep{graves-icml-2014}. The previously best reported basic
seq2seq model on WSJ WER achieved 18.0\% WER \citep{bahdanau-iclr-2016} with
Task Loss Estimation (TLE). Our baseline, also a seq2seq model, achieved
14.8\% WER.  Main differences between our models is that we did not use
convolutional locational-based priors and we used weight noise during training.
The deep CNN model with residual connections, batch normalization and
convolutions achieved a WER of 11.8\% \citep{yu-icassp-2016} \footnote{\label{cnn} For our CNN architectures, we use and compare to the ``(C (3 $\times$ 3) / 2) $\times$ 2 + NiN'' architecture from Table 2 line 4.}.

\begin{table}[t]
  \centering
  \caption{Wall Street Journal test eval92 Word Error Rate (WER) results across Connectionist Temporal Classification (CTC) and Sequence-to-sequence (seq2seq) models. The Latent Sequence Decomposition (LSD) models use a $n=4$ word piece vocabulary (LSD4). The Convolutional Neural Network (CNN) model is with deep residual connections, batch normalization and convolutions. The best end-to-end model is seq2seq + LSD + CNN at 9.6\% WER.}
  \label{tab:compare}
  \begin{tabular}{lr}
    \toprule
    \bfseries Model & \bfseries WER \\
    \midrule
    \cite{graves-icml-2014} \\
    \qquad CTC & 30.1 \\
    \qquad CTC + WER & 27.3 \\
    \midrule
    \cite{hannun-arxiv-2014} \\
    \qquad CTC & 35.8 \\
    \midrule
    \cite{bahdanau-icassp-2016} \\
    \qquad seq2seq & 18.6 \\
    \cite{bahdanau-iclr-2016} \\
    \qquad seq2seq + TLE & 18.0 \\
    \midrule
    \cite{yu-icassp-2016} \\
    \qquad seq2seq + CNN \footnotemark[2] & 11.8 \\
    \midrule
    Our Work \\
    \qquad seq2seq & 14.8 \\
    \qquad seq2seq + LSD4 & 12.9 \\
    \qquad seq2seq + LSD4 + CNN & 9.6 \\
    \bottomrule
  \end{tabular}
\end{table}

Our LSD model using a $n=4$ word piece vocabulary achieves a WER of 12.9\% or
12.7\% relatively better over the baseline seq2seq model. If we combine our LSD
model with the CNN \citep{yu-icassp-2016} model, we achieve a combined WER of
9.6\% WER or 35.1\% relatively better over the baseline seq2seq model. These
numbers are all reported without the use of any language model.

Please see Appendix \ref{appendix} for the decompositions generated by our
model. The LSD model learns multiple word piece decompositions for the same
word sequence.


\section{Related Work}
\cite{singh-ieeetaslp,mccgraw-ieeetaslp,lu-asru-2013} built probabilistic
pronunciation models for Hidden Markov Model (HMM) based systems. However, such
models are still constraint to the conditional independence and Markovian
assumptions of HMM-based systems.

Connectionist Temporal Classification (CTC)
\citep{graves-icml-2006,graves-icml-2014} based models assume conditional
independence, and can rely on dynamic programming for exact inference.
Similarly, \cite{ling-acl-2016} use latent codes to generate text, and also
assume conditional independence and leverage on dynamic programming for exact
maximum likelihood gradients. Such models can not learn the output language if
the language distribution is multimodal. Our seq2seq models makes no such
Markovian assumptions and can learn multimodal output distributions.
\cite{collobert-arxiv-2016} and \cite{zweig-arxiv-2016} developed extensions of
CTC where they used some word pieces. However, the word pieces are only used in
repeated characters and the decompositions are fixed.

Word piece models with seq2seq have also been recently used in machine
translation. \cite{sennirch-acl-2016} used word pieces in rare words, while
\cite{wu-arxiv-2016} used word pieces for all the words, however the
decomposition is fixed and defined by heuristics or another model. The
decompositions in these models are also only a function of the output sequence,
while in LSD the decomposition is a function of both the input and output
  sequence. The LSD framework allows us to learn a distribution of
  decompositions rather than learning just one decomposition defined by a
  priori.

\cite{vinyals-iclr-2016} used seq2seq to outputs sets,
the output sequence is unordered and used fixed length output units; in our
decompositions we maintain ordering use variable lengthed output units.
Reinforcement learning (i.e., REINFORCE and other task loss estimators)
\citep{sutton-1998,graves-icml-2014,ranzato-iclr-2016} learn different output sequences can
yield different task losses. However, these methods don't directly learn different
decompositions of the same sequence. Future work should incorporate LSD with
task loss optimization methods.

\section{Conclusion}
We presented the Latent Sequence Decompositions (LSD) framework. LSD allows us
to learn decompositions of sequences that are a function of both the input and output
sequence. We presented a biased training algorithm based on sampling valid extensions
with an $\epsilon$-greedy strategy, and an approximate decoding algorithm. On the Wall
Street Journal speech recognition task, the sequence-to-sequence character model
baseline achieves 14.8\% WER while the LSD model achieves 12.9\%. Using a 
a deep convolutional neural network on the encoder with LSD, we achieve 9.6\% WER.

\subsubsection*{Acknowledgments}
We thank Ashish Agarwal, Philip Bachman, Dzmitry Bahdanau, Eugene Brevdo, Jan Chorowski, Jeff
Dean, Chris Dyer, Gilbert Leung, Mohammad Norouzi, Noam Shazeer, Xin Pan, Luke
Vilnis, Oriol Vinyals and the Google Brain team for many insightful discussions
and technical assistance.

\bibliography{cites}
\bibliographystyle{iclr2017_conference}

\clearpage
\appendix
\section{Learning the Decompositions}
\label{appendix}
We give the top 8 hypothesis generated by a baseline seq2seq character model, a
Latent Sequence Decompositions (LSD) word piece model and a Maximum Extension
(MaxExt) word piece model. We note that ``shamrock's'' is an out-of-vocabulary
word while ``shamrock'' is in-vocabulary. The ground truth is ``shamrock's
pretax profit from the sale was one hundred twenty five million dollars a
spokeswoman said''. Note how the LSD model generates multiple decompostions for
the same word sequence, this does not happen with the MaxExt model.

\begin{sidewaystable}[h]
  \centering
  \small
  \caption{Top hypothesis comparsion between seq2seq character model, LSD word piece model and MaxExt word piece model.}
  \begin{tabular}{llr}
    \toprule
    $n$ & \bfseries Hypothesis & \bfseries LogProb \\
    \midrule
    \multicolumn{3}{l}{\bfseries Reference} \\
    - & shamrock's pretax profit from the sale was one hundred twenty five million dollars a spokeswoman said & - \\
    \midrule
    \multicolumn{3}{l}{\bfseries Character seq2seq} \\
1 & c|h|a|m|r|o|c|k|'|s| |p|r|e|t|a|x| |p|r|o|f|i|t| |f|r|o|m| |t|h|e| |s|a|l|e| |w|a|s| |o|n|e| |h|u|n|d|r|e|d| |t|w|e|n|t|y| |f|i|v|e| |m|i|l|l|i|o|n| |d|o|l|l|a|r|s| |a| |s|p|o|k|e|s|w|o|m|a|n| |s|a|i|d & -1.373868 \\
2 & c|h|a|m|r|o|x| |p|r|e|t|a|x| |p|r|o|f|i|t| |f|r|o|m| |t|h|e| |s|a|l|e| |w|a|s| |o|n|e| |h|u|n|d|r|e|d| |t|w|e|n|t|y| |f|i|v|e| |m|i|l|l|i|o|n| |d|o|l|l|a|r|s| |a| |s|p|o|k|e|s|w|o|m|a|n| |s|a|i|d & -2.253581 \\
3 & c|h|a|m|r|o|c|k|s| |p|r|e|t|a|x| |p|r|o|f|i|t| |f|r|o|m| |t|h|e| |s|a|l|e| |w|a|s| |o|n|e| |h|u|n|d|r|e|d| |t|w|e|n|t|y| |f|i|v|e| |m|i|l|l|i|o|n| |d|o|l|l|a|r|s| |a| |s|p|o|k|e|s|w|o|m|a|n| |s|a|i|d & -3.482713 \\
4 & c|h|a|m|r|o|c|k|'|s| |p|r|e|t|a|x| |p|r|o|f|i|t| |f|r|o|m| |t|h|e| |s|a|l|e| |w|a|s| |o|n|e| |h|u|n|d|r|e|d| |t|w|e|n|t|y| |f|i|v|e| |m|i|l|l|i|o|n| |d|o|l|l|a|r|s| |o|f| |s|p|o|k|e|s|w|o|m|a|n| |s|a|i|d & -3.493957 \\
5 & c|h|a|m|r|o|d|'|s| |p|r|e|t|a|x| |p|r|o|f|i|t| |f|r|o|m| |t|h|e| |s|a|l|e| |w|a|s| |o|n|e| |h|u|n|d|r|e|d| |t|w|e|n|t|y| |f|i|v|e| |m|i|l|l|i|o|n| |d|o|l|l|a|r|s| |a| |s|p|o|k|e|s|w|o|m|a|n| |s|a|i|d & -3.885185 \\
6 & c|h|a|m|r|o|x| |p|r|e|t|a|x| |p|r|o|f|i|t| |f|r|o|m| |t|h|e| |s|a|l|e| |w|a|s| |o|n|e| |h|u|n|d|r|e|d| |t|w|e|n|t|y| |f|i|v|e| |m|i|l|l|i|o|n| |d|o|l|l|a|r|s| |o|f| |s|p|o|k|e|s|w|o|m|a|n| |s|a|i|d & -4.373687 \\
6 & c|h|a|m|r|o|c|'|s| |p|r|e|t|a|x| |p|r|o|f|i|t| |f|r|o|m| |t|h|e| |s|a|l|e| |w|a|s| |o|n|e| |h|u|n|d|r|e|d| |t|w|e|n|t|y| |f|i|v|e| |m|i|l|l|i|o|n| |d|o|l|l|a|r|s| |a| |s|p|o|k|e|s|w|o|m|a|n| |s|a|i|d & -5.148484 \\
8 & c|h|a|m|r|o|c|k|s| |p|r|e|t|a|x| |p|r|o|f|i|t| |f|r|o|m| |t|h|e| |s|a|l|e| |w|a|s| |o|n|e| |h|u|n|d|r|e|d| |t|w|e|n|t|y| |f|i|v|e| |m|i|l|l|i|o|n| |d|o|l|l|a|r|s| |o|f| |s|p|o|k|e|s|w|o|m|a|n| |s|a|i|d & -5.602793 \\
    \midrule
    \multicolumn{3}{l}{\bfseries Word Piece Model Maximum Extension} \\
1 & sh|am|ro|ck|'s| |pre|ta|x| |pro|fi|t| |from| |the| |sa|le| |was| |one| |hu|nd|red| |tw|ent|y| |five| |mil|lion| |doll|ars| |a| |sp|ok|es|wo|man| |said & -1.155203 \\
2 & sh|am|ro|x| |pre|ta|x| |pro|fi|t| |from| |the| |sa|le| |was| |one| |hu|nd|red| |tw|ent|y| |five| |mil|lion| |doll|ars| |a| |sp|ok|es|wo|man| |said & -3.031330 \\
3 & sh|ar|ro|x| |pre|ta|x| |pro|fi|t| |from| |the| |sa|le| |was| |one| |hu|nd|red| |tw|ent|y| |five| |mil|lion| |doll|ars| |a| |sp|ok|es|wo|man| |said & -3.074762 \\
4 & sh|e| |m| |ro|x| |pre|ta|x| |pro|fi|t| |from| |the| |sa|le| |was| |one| |hu|nd|red| |tw|ent|y| |five| |mil|lion| |doll|ars| |a| |sp|ok|es|wo|man| |said & -3.815662 \\
5 & sh|e| |mar|x| |pre|ta|x| |pro|fi|t| |from| |the| |sa|le| |was| |one| |hu|nd|red| |tw|ent|y| |five| |mil|lion| |doll|ars| |a| |sp|ok|es|wo|man| |said & -3.880760 \\
6 & sh|ar|ro|ck|s| |pre|ta|x| |pro|fi|t| |from| |the| |sa|le| |was| |one| |hu|nd|red| |tw|ent|y| |five| |mil|lion| |doll|ars| |a| |sp|ok|es|wo|man| |said & -4.083274 \\
7 & sh|e| |m| |ro|ck|ed| |pre|ta|x| |pro|fi|t| |from| |the| |sa|le| |was| |one| |hu|nd|red| |tw|ent|y| |five| |mil|lion| |doll|ars| |a| |sp|ok|es|wo|man| |said & -4.878025 \\
8 & sh|e| |m| |ro|ck|s| |pre|ta|x| |pro|fi|t| |from| |the| |sa|le| |was| |one| |hu|nd|red| |tw|ent|y| |five| |mil|lion| |doll|ars| |a| |sp|ok|es|wo|man| |said & -5.121490 \\
    \midrule
    \multicolumn{3}{l}{\bfseries Word Piece Model Latent Sequence Decompositions} \\
1 & sh|a|m|ro|c|k|'s| |pre|ta|x| |pro|fi|t| |fro|m| |t|h|e| |sa|l|e| |was| |on|e| |hu|n|dr|e|d| |t|we|nt|y| |fiv|e| |mil|lio|n| |doll|a|r|s| |a| |sp|ok|e|s|wo|ma|n| |said & -28.111485 \\
2 & sh|a|m|ro|c|k|'s| |pre|ta|x| |pro|fi|t| |fro|m| |t|h|e| |sa|l|e| |was| |on|e| |hu|n|dr|e|d| |t|we|nt|y| |fiv|e| |mil|li|o|n| |doll|ar|s| |a| |sp|ok|e|s|wo|ma|n| |said & -28.172878 \\
3 & sh|a|m|ro|c|k|'s| |pre|ta|x| |pro|fi|t| |fro|m| |t|h|e| |sa|l|e| |was| |on|e| |hu|n|dr|e|d| |t|we|nt|y| |fiv|e| |mil|lio|n| |doll|a|r|s| |a| |sp|ok|e|s|w|om|a|n| |said & -28.453381 \\
4 & sh|a|m|ro|c|k|'s| |pre|ta|x| |pro|fi|t| |fro|m| |t|h|e| |sa|l|e| |was| |on|e| |hu|n|dr|e|d| |t|we|nt|y| |fiv|e| |mil|li|o|n| |doll|a|r|s| |a| |sp|ok|e|s|w|om|a|n| |said & -29.103184 \\
5 & sh|a|m|ro|c|k|'s| |pre|ta|x| |pro|fi|t| |fro|m| |t|h|e| |sa|l|e| |was| |on|e| |hu|n|dr|e|d| |t|we|nt|y| |fiv|e| |mil|lio|n| |doll|a|r|s| |a| |sp|ok|e|s|w|om|a|n| |sa|id & -29.159660 \\
6 & sh|a|m|ro|c|k|'s| |pre|ta|x| |pro|fi|t| |fro|m| |t|h|e| |sa|l|e| |was| |on|e| |hu|n|dr|e|d| |t|we|nt|y| |fiv|e| |mil|lio|n| |doll|a|r|s| |a| |sp|o|k|e|s|w|o|ma|n| |said & -29.164141 \\
7 & sh|a|m|ro|c|k|'s| |pre|ta|x| |pro|fi|t| |fro|m| |t|h|e| |sa|l|e| |was| |on|e| |hu|n|dr|e|d| |t|we|nt|y| |fiv|e| |mil|li|o|n| |doll|a|r|s| |a| |sp|ok|e|s|w|om|a|n| |sai|d & -29.169310 \\
8 & sh|a|m|ro|c|k|'s| |pre|ta|x| |pro|fi|t| |fro|m| |t|h|e| |sa|l|e| |was| |on|e| |hu|n|dr|e|d| |t|we|nt|y| |fiv|e| |mil|li|o|n| |doll|a|r|s| |a| |sp|ok|e|s|w|om|a|n| |sa|id & -29.809937 \\
    \bottomrule
  \end{tabular}
\end{sidewaystable}

\end{document}